\def\paperlanguage{}
\pdfoutput=1 

\documentclass[letterpaper, 10 pt, conference]{ieeeconf}  

\usepackage{bm}
\usepackage{cite}
\usepackage{flushend}
\include{preamble}

\IEEEoverridecommandlockouts                              

\title{\LARGE \textbf
  {
    \switchlanguage%
    {%
      Adaptive Body Schema Learning System Considering Additional Muscles for Musculoskeletal Humanoids
    }%
    {%
      筋骨格ヒューマノイドにおける筋増加を考慮可能な適応的身体図式学習システム
    }%
  }
}

\author{Kento Kawaharazuka$^{1}$, Akihiro Miki$^{1}$, Yasunori Toshimitsu$^{1}$, Kei Okada$^{1}$, and Masayuki Inaba$^{1}$
  \thanks{$^{1}$ The authors are with the Department of Mechano-Informatics, Graduate School of Information Science and Technology, The University of Tokyo, 7-3-1 Hongo, Bunkyo-ku, Tokyo, 113-8656, Japan.
    {\texttt\small [kawaharazuka, miki, toshimitsu, k-okada, inaba]@jsk.t.u-tokyo.ac.jp}
  }
}
\begin{document}

\maketitle
\thispagestyle{empty}
\pagestyle{empty}

\begin{abstract}
  \switchlanguage%
  {%
    One of the important advantages of musculoskeletal humanoids is that the muscle arrangement can be easily changed and the number of muscles can be increased according to the situation.
    In this study, we describe an overall system of muscle addition for musculoskeletal humanoids and the adaptive body schema learning while taking into account the additional muscles.
    For hardware, we describe a modular body design that can be fitted with additional muscles, and for software, we describe a method that can learn the changes in body schema associated with additional muscles from a small amount of motion data.
    We apply our method to a simple 1-DOF tendon-driven robot simulation and the arm of the musculoskeletal humanoid Musashi, and show the effectiveness of muscle tension relaxation by adding muscles for a high-load task.
  }%
  {%
    筋骨格ヒューマノイドの重要な利点として, 筋配置の変更が容易であり, 状況に応じて筋を増やすことができる点が挙げられる.
    本研究では, 筋骨格ヒューマノイドの筋増加とそれに伴う適応的身体図式学習の全体システムについて述べる.
    ハードウェアとして, 筋を追加で装着できるモジュラー型の身体設計, ソフトウェアとして, 筋の増加に伴う身体図式の変化を少ない追加データから再学習可能な手法について述べる.
    1自由度腱駆動シミュレーション, 筋骨格ヒューマノイドMusashiの腕について本手法を適用し, 高負荷タスクへの筋追加による筋張力緩和の有効性を示す.
  }%
\end{abstract}

\section{INTRODUCTION}\label{sec:introduction}
\switchlanguage%
{%
  A variety of human mimetic musculoskeletal humanoids have been developed so far \cite{gravato2010ecce1, asano2016kengoro, kawaharazuka2019musashi}.
  Since they mimic the human body in detail, they have various biomimetic advantages such as body flexibility, redundant muscles, ball joints and the spine \cite{mizuuchi2005reinforceablemuscle}.

  Among these, redundant muscles are one of the most important features.
  First, this enables variable stiffness control using antagonistic muscles and nonlinear elastic elements, which has been used for environmental contact and movements with impact \cite{kobayashi1998tendon, kawaharazuka2019longtime, nakanishi2011kenzoh}.
  In addition, a robust motion strategy using redundant muscles, in which the robot can continue to move even if one muscle is broken \cite{kawamura2016jointspace}, and a design optimization method maximizing the redundancy \cite{kawaharazuka2021redundancy} have been developed.
  It is also important to note that it is possible to easily increase the number of muscles or change the muscle arrangement depending on the task, and this has been used to realize a stable standing posture \cite{mizuuchi2006reinforcing} and to optimize the muscle arrangement depending on the task \cite{nakanishi2006arrangement}.
  On the other hand, high internal muscle tension sometimes occurs due to the existence of antagonistic muscles and model errors.
  To solve these problems, antagonist inhibition control \cite{kawaharazuka2017antagonist} and muscle relaxation control \cite{kawaharazuka2019relax} have been proposed.
  In addition, since the maximum joint velocity is limited to the slowest muscle among the redundant muscles, a method to solve this problem has also been developed \cite{kawaharazuka2020speed}.
}%
{%
  これまで様々な人体模倣型の筋骨格ヒューマノイドが開発されたきた\cite{gravato2010ecce1, asano2016kengoro, kawaharazuka2019musashi}.
  これらは, 人体を詳細に模倣しているがゆえに, その柔軟身体, 冗長な筋肉, 可変剛性制御\cite{kobayashi1998tendon}, 球関節や背骨\cite{mizuuchi2005reinforceablemuscle}等様々な生物規範型の利点を有する.

  その中でも冗長な筋肉は最も重要な特徴の一つである.
  この利点としてまず, 主動筋拮抗筋による冗長性と非線形弾性要素を使った可変剛性制御が挙げられ, 環境接触行動や衝撃を伴う動作への利用が行われている\cite{kobayashi1998tendon, kawaharazuka2019longtime, nakanishi2011kenzoh}.
  また, 筋が一本切れても動作し続けることが可能な冗長駆動系を利用したロバストな動作戦略\cite{kawamura2016jointspace}や, この筋破断時の冗長性を最大化する設計最適化手法\cite{kawaharazuka2021redundancy}が開発されている.
  また, タスクに応じて容易に筋を増やしたり筋配置を変更したりすることが可能な点も重要であり, これを使った立位姿勢の実現\cite{mizuuchi2006reinforcing}やタスクに応じた筋配置の最適化が行われている\cite{nakanishi2006arrangement}.
  一方欠点として, 主動筋と拮抗筋の存在とモデル誤差による筋内力の高まりが挙げられ, これらを解決する拮抗筋抑制制御\cite{kawaharazuka2017antagonist}, 筋弛緩制御\cite{kawaharazuka2019relax}, 拮抗筋修正制御\cite{koga2019modification}等が提案されている.
  また, 冗長な筋の中でも最も遅い筋肉に最大関節速度が制限されてしまうため, これを解決する手法も開発されている\cite{kawaharazuka2020speed}.
}%

\begin{figure}[t]
  \centering
  \includegraphics[width=1.0\columnwidth]{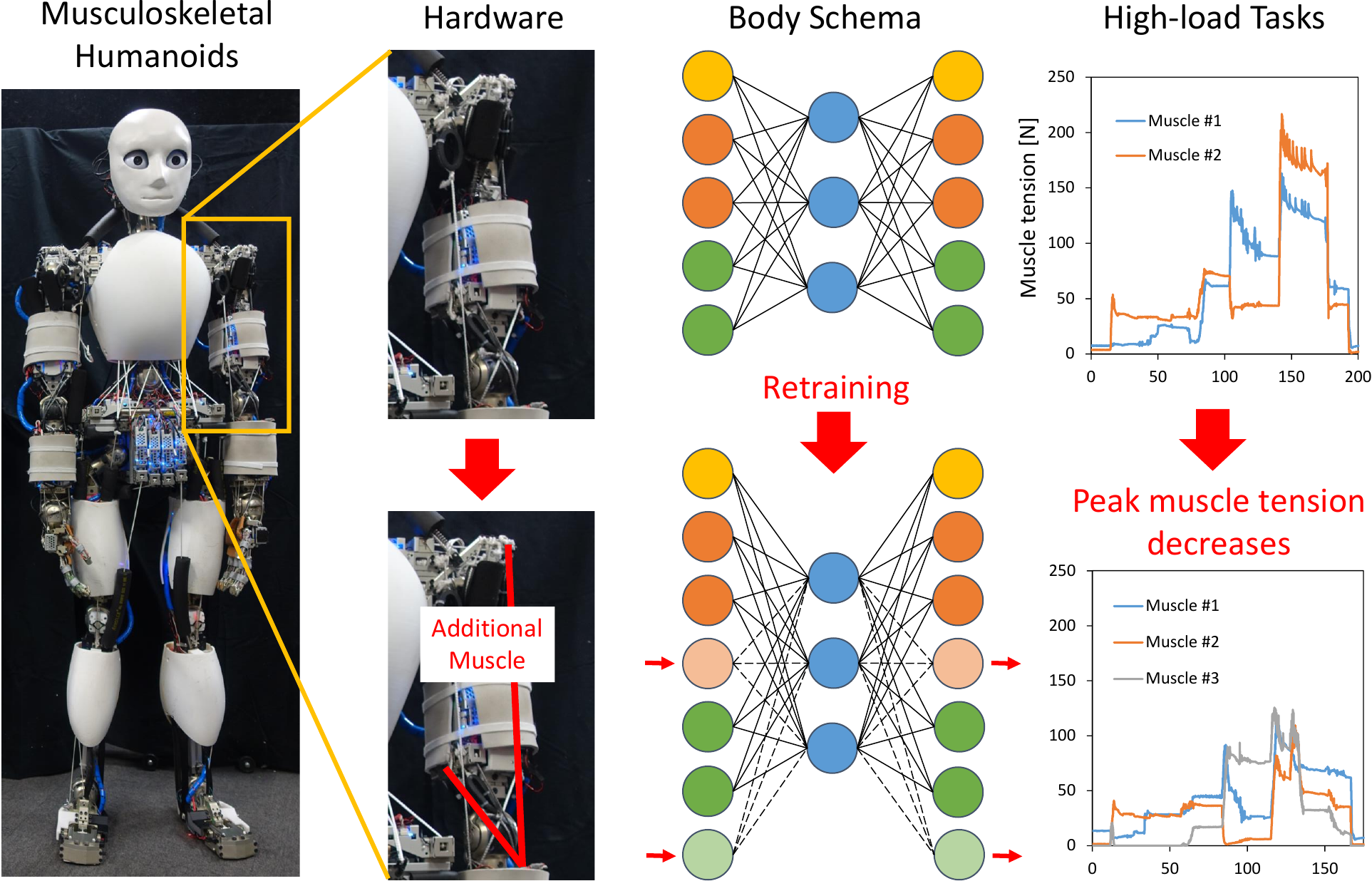}
  \vspace{-3.0ex}
  \caption{The concept of the entire system of adaptive body schema learning considering muscle addition.}
  \label{figure:concept}
  \vspace{-1.0ex}
\end{figure}

\begin{figure}[t]
  \centering
  \includegraphics[width=0.6\columnwidth]{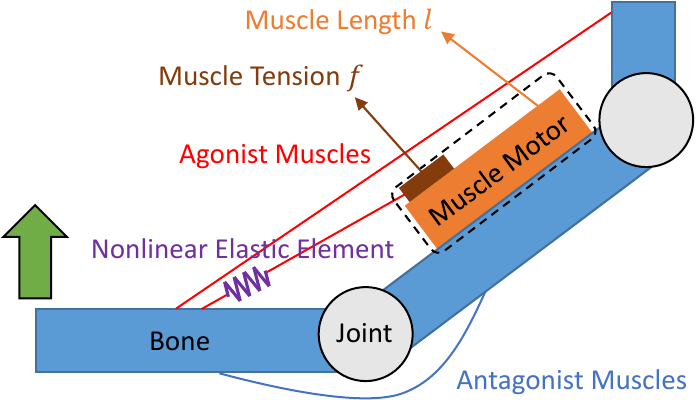}
  \vspace{-1.0ex}
  \caption{The basic musculoskeletal structure.}
  \label{figure:basic-structure}
  \vspace{-3.0ex}
\end{figure}

\begin{figure*}[t]
  \centering
  \includegraphics[width=1.9\columnwidth]{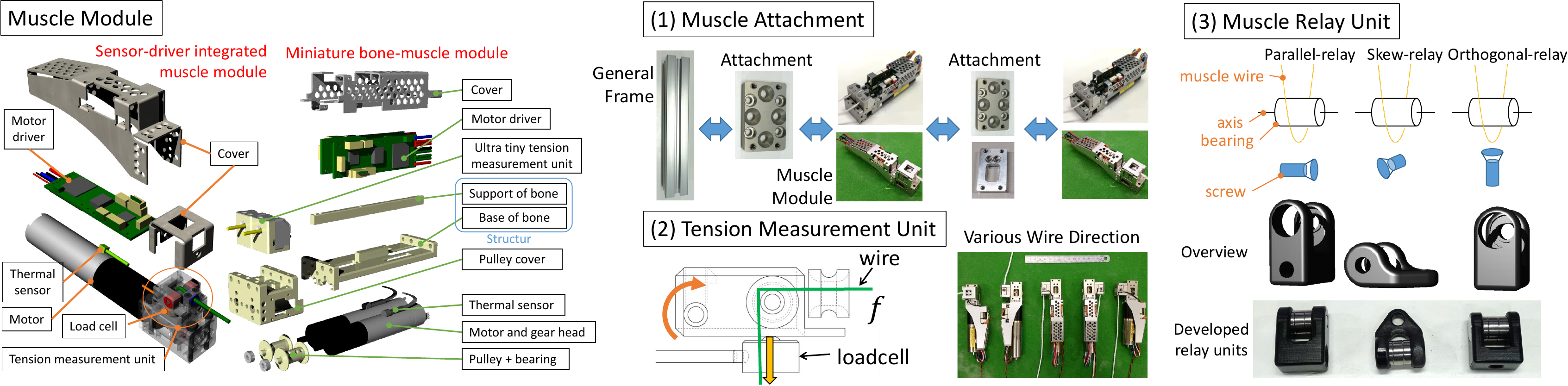}
  \vspace{-1.0ex}
  \caption{The hardware design considering muscle addition.}
  \label{figure:additional-muscles}
  \vspace{-3.0ex}
\end{figure*}

\switchlanguage%
{%
  In this study, we focus on a task-dependent muscle addition; that is, the addition of sensors and actuators in the body.
  Task-dependent muscle addition is a very attractive feature not found in axis-driven robots.
  In previous studies, muscle modules have been added in the middle of the muscle wire depending on the task, and they dangled in midair \cite{mizuuchi2006reinforcing}.
  However, the circuit wiring of the muscle module also floats in midair, making it unreliable.
  Thus, the method of attaching the muscle modules directly to the skeleton is becoming more common \cite{jantsch2013anthrob, kawaharazuka2019musashi}.
  Therefore, in this study, we will increase the number of muscles by using attachments that connect muscle modules to each other.
  The direction of the muscles can be freely changed depending on the attached direction of the muscle tension measurement unit, and free muscle placement can be realized by the standardized muscle relay units.
  We reconsider the body structure of the developed musculoskeletal humanoid Musashi \cite{kawaharazuka2019musashi} from the viewpoint of the muscle addition.

  Also, in previous studies, humans have modelized the moment arm and muscle arrangement after adding muscles, and generated the movements manually.
  In this study, we use a body schema model \cite{kawaharazuka2020autoencoder}, which represents the relationship among joint angle, muscle tension, and muscle length, to control the body of the musculoskeletal humanoid.
  We then discuss a method to relearn the body schema model changed by the addition of muscles using a small amount of motion data (\figref{figure:concept}).
  In other words, after the addition of muscles, the system can automatically acquire motion data, relearn the body schema, and resume the movement.
  Note that since force control is difficult for musculoskeletal humanoids due to friction issues \cite{kawamura2016jointspace}, position control of muscle length is used in this study.

  We introduce some previous studies regarding body schema learning.
  So far, learning methods of the mapping between joint angle and muscle length \cite{mizuuchi2006acquisition, kawaharazuka2018online}, the mapping between muscle length and operational position \cite{motegi2012jacobian}, the mapping among joint angle, muscle tension, and muscle length \cite{kawaharazuka2018bodyimage, kawaharazuka2019longtime}, etc., have been proposed.
  These methods modelize the complex and flexible musculoskeletal structures using neural networks, and perform control or state estimation.
  In this study, we use Musculoskeletal AutoEncoder (MAE) \cite{kawaharazuka2020autoencoder}, which enables control, simulation, state estimation, and anomaly detection in a single network, which have been constructed separately in the past.
  This means that the change in only this single network needs to be considered for muscle addition.
  Here, it is important to note that not only MAE but also most networks \cite{kawaharazuka2018bodyimage, kawaharazuka2019longtime} include muscle sensor information such as muscle length and muscle tension in both the input and output of the network.
  Therefore, it is necessary to take into consideration the increase in the input and output dimensions when using any network, and this study can be applied to networks other than MAE.
  On the other hand, when considering learning systems for robots other than musculoskeletal robots, studies dealing with the increase of the output dimension can be found in the context of incremental learning.
  While preventing catastrophic forgetting, the output dimension of the network is increased, and the network is continuously updated with new data \cite{li2018forgetting, masana2020incremental}.
  However, changes in the input dimension have rarely been addressed.
  In addition, most of the tasks are image recognition tasks where the number of labels to be classified increases, and there are no applications to regression problems on robot sensors and actuators.
  This is because robots are assumed to be systems whose sensors and actuators do not change or grow in most cases.
  This study is technically new in that it deals with changes in the input dimension as well as the output dimension of the network, and solves the regression problem of sensors and actuators.

  This study describes the development of an adaptive body schema learning system for musculoskeletal humanoids that can easily add new muscles.
  The contribution is as follows.
  \begin{itemize}
    \item Requirements and design of hardware considering muscle addition in musculoskeletal humanoids
    \item Relearning of body schema with a small amount of motion data considering additional muscles
    \item Task realization with an adaptive body schema learning system considering muscle addition
  \end{itemize}

}%
{%
  本研究はこれらの中でも, タスクに応じた筋の増加, つまり身体のセンサ・アクチュエータの増加について扱う.
  タスクに応じた筋の増加は軸駆動型ロボットにはない非常に魅力的な特徴である.
  これまでの研究では, 筋ワイヤの中間に筋アクチュエータを追加し, 筋がぶらぶらと宙吊りになる構造の筋モジュールをタスクに応じて追加していた\cite{mizuuchi2006reinforcing}.
  しかし, 筋と同時にその配線も中に浮くため, 信頼性に問題があり, 骨格に直接筋モジュールを貼り付ける方式が主流になりつつある\cite{jantsch2013anthrob, kawaharazuka2019musashi}.
  そのため, 本研究では筋モジュールと筋モジュール間を繋ぐアタッチメントにより, 筋を増加させていく.
  筋張力測定ユニットの取り付け方向次第で筋の方向を自在に変化させ, 筋経由点ユニットにより自由な筋配置を実現する.
  これらの筋追加の観点から, 開発した筋骨格ヒューマノイドMusashi \cite{kawaharazuka2019musashi}の身体構造を捉え直す.

  また, これまで筋を追加した後は人間がそのモーメントアームや筋配置をモデル化し, 手作業で動作を生成していた.
  本研究では, 関節角度-筋張力-筋長の相互関係を表現する身体図式\cite{kawaharazuka2020autoencoder}を筋骨格ヒューマノイドの制御に用い, 筋の増加により変化した身体図式ネットワークを少数のデータから学習し直す手法について考察する(\figref{figure:concept}).
  つまり, 筋の追加後, 自動でデータを取得, 身体図式を再学習し, 動作を再開することが可能になる.
  なお, 筋については力制御でも位置制御でも, モーメントアームが分からないため再学習が必要となるが, 現状筋骨格ヒューマノイドは摩擦の問題から力制御が難しいため\cite{kawamura2016jointspace}, 本研究では位置制御を利用している.

  身体図式学習の観点で, いくつかの先行研究を紹介する.
  これまで, 関節角度-筋長マッピング学習\cite{mizuuchi2006acquisition, kawaharazuka2018online}, 作業空間-筋長マッピング学習\cite{motegi2012jacobian}, 関節角度-筋張力-筋長マッピング\cite{kawaharazuka2018bodyimage, kawaharazuka2019longtime}等, 様々な身体図式学習が提案されてきた.
  これらは, 複雑で柔軟な筋骨格構造をニューラルネットワークによりモデル化し, 制御や状態推定を行う.
  その中でも本研究では, これまでバラバラに開発されてきた制御やシミュレーション, 状態推定, 異常検知等を一つのネットワークで可能にしたMusculoskeletal AutoEncoderを用いる\cite{kawaharazuka2020autoencoder}.
  これにより, たった一つのネットワークについてのみ, 筋の増加を考えれば良い.
  ここで重要なのは, Musculoskeletal AutoEncoderに限らず, ほとんどのネットワーク\cite{kawaharazuka2018bodyimage, kawaharazuka2019longtime}において, ネットワークの入力にも出力にも筋長や筋張力等の筋センサ情報が入るため, 入力次元の増加と出力次元の増加を扱う必要があるという点である.
  一方, 筋骨格だけではない学習システムについて考えると, 出力次元の増加を扱った研究はincremental learningの文脈で見られる.
  破壊的忘却を防ぎながらも, ネットワークの出力を増やし, 新しいデータにより継続的に学習を行う\cite{li2018forgetting, masana2020incremental}.
  一方, 入力次元の変化についてはほとんど扱われてきていない.
  また, 大抵は画像認識タスクについてその分類するラベルが増えるというタスクであり, ロボットのセンサ・アクチュエータに関する回帰問題への応用はない.
  これは, ほとんどの場合ロボットはそのセンサ・アクチュエータが変化しない, 成長しないシステムであることを前提としているからであると考えられる.
  本研究はネットワークの出力次元だけでなく入力次元の変化まで扱い, かつセンサ・アクチュエータの回帰問題を解決する点で技術的に新しい.

  これらの観点から, 本研究では筋を容易に追加可能な筋骨格ヒューマノイドにおける, 筋増加を考慮可能な身体図式学習システムの開発について述べる.
  本研究のcontributionは以下である.
  \begin{itemize}
    \item 筋骨格ヒューマノイドにおける筋増加に対応したハードウェアの要件と設計
    \item 筋増加に伴う身体図式変化時における少数のデータによる身体図式の再学習
    \item 筋増加を考慮した身体図式学習システムによるタスク実現
  \end{itemize}

  本研究の構成は以下である.
  \secref{sec:musculoskeletal-humanoids}では, 筋骨格ヒューマノイドの特徴と容易に筋を増やすことのできる身体構成について述べる.
  \secref{sec:proposed}では, 筋が増えた際のネットワークの再学習方法について, Musculoskeletal AutoEncoderの詳細, データ取得方法, ネットワークの重みコピー, 学習方法を述べる.
  \secref{sec:experiments}では, 1自由度腱駆動シミュレーションと筋骨格ヒューマノイドMusashiの筋増加に伴う身体図式学習実験, これによる高負荷タスクの実現について述べる.
  \secref{sec:discussion}では実験結果について議論し, \secref{sec:conclusion}で結論を述べる.
}%

\begin{figure*}[t]
  \centering
  \includegraphics[width=1.9\columnwidth]{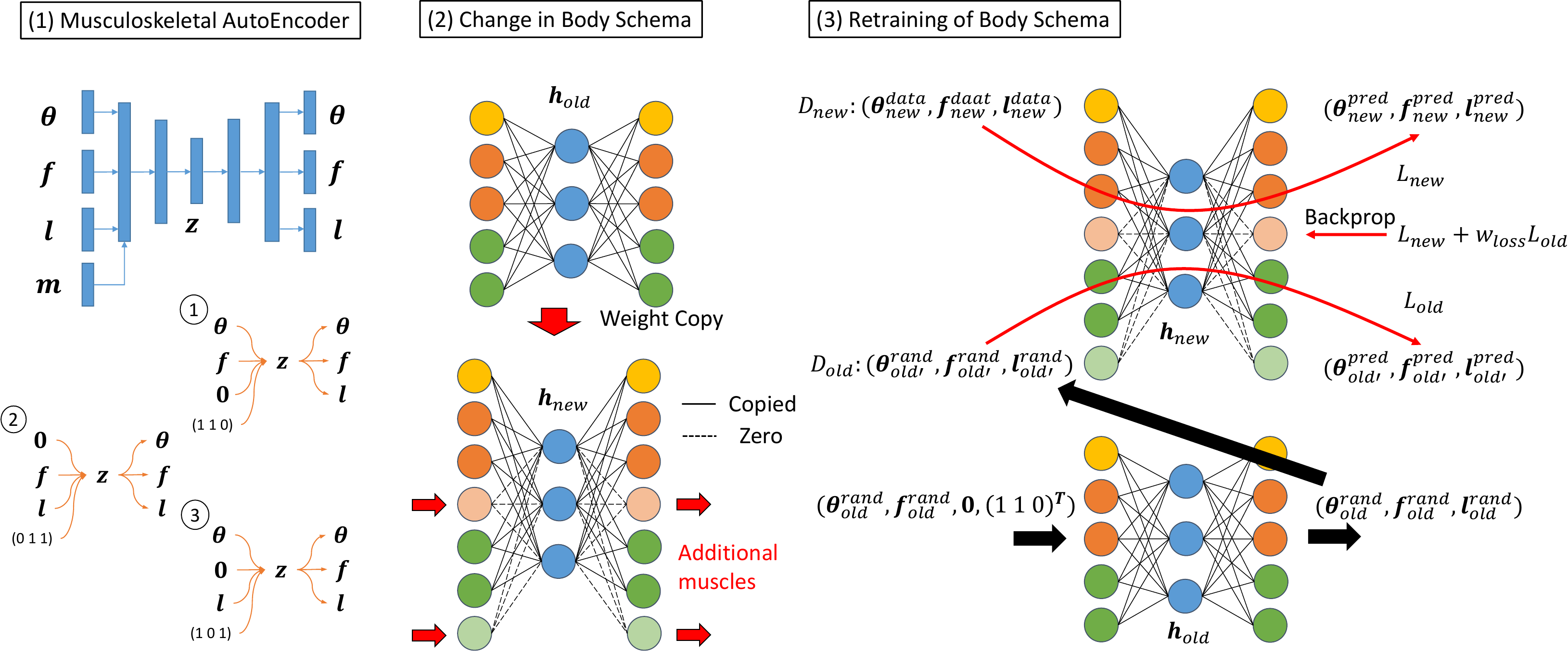}
  \vspace{-1.0ex}
  \caption{The adaptive body schema learning system considering additional muscles.}
  \label{figure:system-overview}
  \vspace{-3.0ex}
\end{figure*}

\section{Structure of Musculoskeletal Humanoids Considering Muscle Addition} \label{sec:musculoskeletal-humanoids}
\subsection{Basic Structure of Musculoskeletal Humanoids} \label{subsec:basic-structure}
\switchlanguage%
{%
  The basic musculoskeletal structure is shown in \figref{figure:basic-structure}.
  Redundant muscles are arranged antagonistically around the joints.
  The muscles are mainly composed of an abrasion resistant synthetic fiber Dyneema, and nonlinear elastic elements that allow variable stiffness control are often placed in series with the muscles.
  In some robots, the muscles are folded back using pulleys to increase the moment arm.
  In some cases, the muscle is covered by a soft foam material for flexible contact, making the modeling of the robot more difficult.
  For each muscle, muscle length $l$, muscle tension $f$, and muscle temperature $c$ can be measured.
  The joint angle $\bm{\theta}$ cannot be measured in many cases due to ball joints and the complex scapula, but it can be measured using special mechanisms in some robots \cite{kawaharazuka2019musashi}.
  Even in cases where its direct measurement is not possible, the joint angle of the actual robot can be estimated by using markers attached to the hand, joint angle estimation based on muscle length changes, and inverse kinematics \cite{kawaharazuka2018online}.
}%
{%
  基本的な筋骨格構造を\figref{figure:basic-structure}に示す.
  冗長な筋が関節の周りに拮抗して配置されている.
  筋は主に摩擦に強い合成繊維であるDyneemaによって構成されており, 可変剛性を可能とする非線形性弾性要素が筋と直列に配置されている場合が多い.
  ロボットによっては, モーメントアームを稼ぐために動滑車を使って筋を折り返している場合も存在する.
  柔軟な接触のために, 筋の周りには外装としての発泡材が巻かれている場合もあり, モデル化は困難である.
  それぞれの筋について筋長$l$・筋張力$f$・筋温度$c$が測定できる.
  関節角度$\bm{\theta}$は球関節や複雑な肩甲骨ゆえに測定できない場合が多いが, 一部のロボットでは測定することが可能である\cite{kawaharazuka2019musashi}.
  直接測定できない場合も, 手につけたマーカ, 筋長変化による関節角度推定, 逆運動学の併用により実機の関節角度を推定することが可能である\cite{kawaharazuka2018online}.
}%

\subsection{Hardware Design Considering Muscle Addition} \label{subsec:addtional-muscles}
\switchlanguage%
{%
  The structure of the musculoskeletal humanoid Musashi is shown in \figref{figure:additional-muscles}.
  The muscles are composed of muscle modules \cite{asano2015sensordriver, kawaharazuka2017forearm} that integrate actuators, pulleys, motor drivers, muscle tension measurement units, etc., as shown in the left figure of \figref{figure:additional-muscles}.
  This increases the reliability and modularity, and facilitates muscle replacement and muscle addition.
  Here, we consider that the following three points are necessary for a body structure that allows muscle addition.
  \begin{enumerate}
    \renewcommand{\labelenumi}{(\arabic{enumi})}
    \item The muscle modules can be attached to various parts of the body.
    \item The muscle wires can be routed from the muscle module in various directions.
    \item Arbitrary muscle paths can be realized by specifying various muscle relay points.
  \end{enumerate}
  These points allow us to add muscles at arbitrary locations in arbitrary muscle paths.
  (1) is made possible by muscle attachment, as shown in (1) of \figref{figure:additional-muscles}.
  The skeleton is composed of generic aluminum frames, and muscle modules are connected to it by muscle attachment.
  Also, muscle modules can be connected to each other by the same muscle attachment.
  (2) is made possible by the muscle tension measurement unit, which can realize various muscle wire directions, as shown in (2) of \figref{figure:additional-muscles}.
  The muscle tension measurement unit measures the muscle tension as the moment of directional change in the muscle wire pushes the loadcell.
  This unit can be connected to the four sides of the muscle module and the muscle can go be routed from the unit in four directions.
  (3) is made possible by muscle relay units that can realize various muscle paths, as shown in (3) of \figref{figure:additional-muscles}.
  This unit can be standardized based on the combination of the muscle wire direction and the direction of the muscle relay unit attached to the skeleton, and arbitrary muscle paths can be realized by these combinations.

  The circuit configuration is also briefly described.
  The entire circuit is connected by USB communication.
  The motor drivers in the muscle modules are connected to each other by daisy chain from a USB HUB board located in each region.
  To add a new muscle, we only need to extend the cable from the nearby muscle.

  Note that it took a skilled researcher about three minutes to install an additional muscle module, muscle attachment, muscle wires, and cables.
}%
{%
  筋骨格ヒューマノイドMusashiの筋骨格構造について\figref{figure:additional-muscles}に示す.
  筋は, \figref{figure:additional-muscles}の左図に示すように, アクチュエータ・プーリ・モータドライバ・筋張力測定ユニット等を一体化した筋モジュールにより構成されている\cite{asano2015sensordriver, kawaharazuka2017forearm}.
  これにより, 信頼性やモジュール性が上がり, 筋の交換や筋の追加が容易になる.
  ここで, 筋追加を可能な身体構造に必要な点は以下の3つである.
  \begin{enumerate}
    \renewcommand{\labelenumi}{(\arabic{enumi})}
    \item 身体の様々な部位に筋モジュールを取り付けることができる.
    \item 筋ワイヤを筋モジュールから様々な方向に出すことができる.
    \item 筋の経由点を様々に指定し, 任意の筋経路を実現できる.
  \end{enumerate}
  これらにより, 任意の筋経路で, 任意の場所に筋を追加することができるようになる.
  (1)は, \figref{figure:additional-muscles}の(1)にあるように, Muscle Attachmentにより可能になる.
  全身の骨格を汎用フレームとし, これと筋モジュールをMuscle Attachmentにより接続する.
  また, 筋モジュールと筋モジュール同士もMuscle Attachmentにより接続することができる.
  (2)は, \figref{figure:additional-muscles}の(2)にあるように, 様々な筋ワイヤ方向を実現可能な筋張力測定ユニットにより可能になる.
  筋張力測定ユニットは筋ワイヤの方向転換におけるモーメントによりロードセルを押しこむことで筋張力を測定できる.
  このユニットが筋モジュールの四面に対して接続可能であり, かつ, そこから四方向全てに出て行くことができる.
  (3)は, \figref{figure:additional-muscles}の(3)にあるように, 様々な筋経路を実現可能な筋経路ユニットにより可能になる.
  筋ワイヤが出る方向と筋経路ユニットを骨格に取り付ける方向の組み合わせからこのユニットを標準化し, これらの組み合わせにより任意の筋経路を実現することができる.

  回路構成についても簡単に触れる.
  回路全体はUSB通信によって通信している.
  各部位に配置されたUSBハブ基板からデイジーチェーンで筋モジュール内のモータドライバ同士が接続されている.
  新しい筋を追加する際は, 近くに配置されている筋からケーブルを伸ばすのみで良い.

  なお, Muscle Attachment, 筋モジュール, 筋ワイヤ, ケーブル類を新しく取り付けるのを, 熟練した研究者は約3分で終えることができた.
}%

\section{Adaptive Body Schema Learning System Considering Muscle Addition} \label{sec:proposed}
\switchlanguage%
{%
  The overall structure of this system is shown in \figref{figure:system-overview}.
}%
{%
  本システムの全体構成を\figref{figure:system-overview}に示す.
  Musculoskeletal AutoEncoderの詳細構造, 筋増加の際のネットワーク構造変化, ネットワークの再学習手法が順に示されている.
}%
\subsection{Body Schema Learning: Musculoskeletal AutoEncoder}
\switchlanguage%
{%
  First, we describe the body schema model used in this study, Musculoskeletal AutoEncoder (MAE) \cite{kawaharazuka2020autoencoder}.
  MAE is a single network that represents the three relations among $(\bm{\theta}, \bm{f}, \bm{l})$: $(\bm{\theta}, \bm{f})\to\bm{l}$, $(\bm{f}, \bm{l})\to\bm{\theta}$, and $(\bm{l}, \bm{\theta})\to\bm{f}$.
  An AutoEncoder-type network $\bm{h}$ with $(\bm{\theta}, \bm{f}, \bm{l})$ and a mask value $\bm{m}$ as input, $\bm{z}$ as the latent space, and $(\bm{\theta}, \bm{f}, \bm{l})$ as output, is updated from the actual robot sensor information.
  The mask $\bm{m}$ has three values: $\begin{pmatrix}1&1&0\end{pmatrix}^{T}$, $\begin{pmatrix}0&1&1\end{pmatrix}^{T}$, and $\begin{pmatrix}1&0&1\end{pmatrix}^{T }$.
  For example, if $\bm{m}=\begin{pmatrix}1&1&0\end{pmatrix}^{T}$, we take $(\bm{\theta}, \bm{f}, \bm{0}, \begin{pmatrix}1&1&0\end{pmatrix}^{T })$ as input, and MAE outputs $(\bm{\theta}, \bm{f}, \bm{l})$ through $\bm{h}$.
  Here, to calculate the current estimated joint angle $\bm{\theta}^{est}$ from the information of $(\bm{f}, \bm{l})$, we can use the mask $\bm{m}=\begin{pmatrix}0&1&1\end{pmatrix}^{T}$.
  Also, to calculate the target muscle length $\bm{l}^{ref}$ to achieve the target joint angle $\bm{\theta}^{ref}$, we calculate $\bm{z}$ from $\bm{\theta}^{ref}$ and the appropriate $\bm{f}$.
  Then, $(\bm{\theta}, \bm{f}, \bm{l})$ is output from $\bm{z}$, and for this value, we calculate the loss considering the constraints that $\bm{\theta}$ approaches $\bm{\theta}^{ref}$, minimizes $\bm{f}$, and exerts the required joint torque.
  Based on this loss, $\bm{z}$ can be iteratively updated by back propagation and gradient descent method to finally calculate the target muscle length $\bm{l}^{ref}$.
  Note that MAE represents only the static intersensory relationship, so it cannot absorb model errors due to friction, hysteresis, etc.
}%
{%
  まず, 本研究で扱う身体図式モデル, Musculoskeletal AutoEncoder (MAE) \cite{kawaharazuka2020autoencoder}について述べる.
  MAEは, $(\bm{\theta}, \bm{f}, \bm{l})$の間の関係である, $(\bm{\theta}, \bm{f})\to\bm{l}$, $(\bm{f}, \bm{l})\to\bm{\theta}$, $(\bm{l}, \bm{\theta})\to\bm{f}$という3つの関係を一つのネットワークで表したものである.
  入力を$(\bm{\theta}, \bm{f}, \bm{l})$とマスク値$\bm{m}$として, 潜在空間$\bm{z}$を通して,  出力を$(\bm{\theta}, \bm{f}, \bm{l})$としたAutoEncoder型のネットワーク$\bm{h}$を実機センサデータから更新する.
  マスク$\bm{m}$は, $\begin{pmatrix}1&1&0\end{pmatrix}^{T}$, $\begin{pmatrix}0&1&1\end{pmatrix}^{T}$, $\begin{pmatrix}1&0&1\end{pmatrix}^{T}$の3つの値を取り, 例えば$\bm{m}=\begin{pmatrix}1&1&0\end{pmatrix}^{T}$のとき, $(\bm{\theta}, \bm{f}, \bm{0}, \begin{pmatrix}1&1&0\end{pmatrix}^{T})$を入力として, $(\bm{\theta}, \bm{f}, \bm{l})$を出力する.
  ここで, 例えば現在の推定関節角度$\bm{\theta}^{est}$を$(\bm{f}, \bm{l})$の情報から計算するためには, マスク$\bm{m}=\begin{pmatrix}0&1&1\end{pmatrix}^{T}$を用いれば良い.
  また, ある指令値$\bm{\theta}^{ref}$を実現するための筋長$\bm{l}^{ref}$を計算するためには, $\bm{\theta}^{ref}$と適当な$\bm{f}$から$\bm{z}$を計算する.
  そして, $\bm{z}$から$(\bm{\theta}, \bm{f}, \bm{l})$を出力し, この値に対して, $\bm{\theta}$が$\bm{\theta}^{ref}$に近づく制約, $\bm{f}$を最小化する制約, 必要な関節トルクを発揮する制約を考慮した損失を計算する.
  この損失をもとに, $\bm{z}$を誤差逆伝播と勾配法により繰り返し更新することで, 最終的に指令筋長$\bm{l}^{ref}$を計算することができる.
  なお, MAEは静的なセンサ関係のみ表しているため, 摩擦やヒステリシス等によるモデル誤差までは吸収できない.
}%

\subsection{Change in Body Schema by Muscle Addition} \label{subsec:changed-schema}
\switchlanguage%
{%
  The model of MAE, $\bm{h}$, changes with muscle addition.
  We call the model before muscle addition $\bm{h}_{old}$ and the model after the muscle addition $\bm{h}_{new}$.
  The number of joints used for $\bm{h}$ is $N$, the number of muscles is $M$, and the number of muscles before and after muscle addition is $M_{\{old, new\}}$ ($M_{new}>M_{old}$).
  The dimension of input and output of MAE changes from ($N, M_{old}, M_{old}$) to ($N, M_{new}, M_{new}$).
  Since it is inefficient to completely learn the model of $\bm{h}_{new}$ from scratch, we copy the network weight of $\bm{h}_{old}$ to $\bm{h}_{new}$ as shown in (2) of \figref{figure:system-overview}.
  For the rest of the model, we set both the weights and the bias of the network to zero.
  As a result, before the relearning of $\bm{h}_{new}$, no matter what values are put into $\bm{f}$ and $\bm{l}$ for the additional muscles in the input, the values of the original muscles show the same behavior as $\bm{h}_{old}$.
  Note that MAE is actually constructed as a five-layered network, but in (2) and (3) of \figref{figure:system-overview}, it is abbreviated to a 3-layered network in visual.
}%
{%
  筋増加により, MAEのモデルである$\bm{h}$は変化する.
  筋増加前のモデルを$\bm{h}_{old}$, 筋増加後のモデルを$\bm{h}_{new}$とする.
  $\bm{h}$に用いられる関節数を$N$, 筋数を$M$とし, 筋増加前と後の筋数を$M_{\{old, new\}}$とする($M_{new}>M_{old}$).
  MAEの入出力の次元は($N, M_{old}, M_{old}$)から($N, M_{new}, M_{new}$)へと変化する.
  ここで, $\bm{h}_{new}$のモデルを完全に最初から学習し直すのは非常に効率が悪いため, $\bm{h}_{old}$の重みをコピーして用いる.
  コピーされる以外の部分については, ネットワークの重み・バイアスをともに0とする.
  これにより, $\bm{h}_{new}$の学習前については, 入力において追加された筋の$\bm{f}$, $\bm{l}$にどんな値を入れても, 元々ある筋については$\bm{h}_{old}$と同様の挙動を示す.
  なお, 実際にはMAEは5層のネットワークで記述されているが, \figref{figure:system-overview}の(2), (3)においては3層のネットワークに省略して記述している.
}%

\subsection{Data Collection for Body Schema Learning} \label{subsec:data-collection}
\switchlanguage%
{%
  First, each muscle is operated by the following muscle stiffness control \cite{shirai2011stiffness}.
  \begin{align}
    \bm{f}^{ref} = \bm{f}_{bias} + k_{stiff}(\bm{l}-\bm{l}^{ref}) \label{eq:msc}
  \end{align}
  $\bm{f}_{bias}$ is the bias term of the muscle stiffness control, and $k_{stiff}$ is the muscle stiffness coefficient.

  After increasing the number of muscles, it is necessary to obtain new motion data.
  In this case, since the relationship among $(\bm{\theta}, \bm{f}, \bm{l})$ is not known for the newly added muscle, we collect motion data for the additional muscle by actuating it differently from the other muscles.
  Here, $k_{stiff}$ in \equref{eq:msc} is set to 0 only for the newly added muscle and it does not follow the target muscle length.
  Instead, $\bm{f}^{bias}$ is specified randomly.
  For existing muscles, we input $(\bm{\theta}^{rand}_{new}, \bm{f}^{rand}_{new}, \bm{0}, \begin{pmatrix}1&1&0\end{pmatrix}^{T})$ to $\bm{h}_{new}$, and send the obtained $\bm{l}^{ref}$ to the actual robot ($\{\bm{\theta}, \bm{f}\}^{rand}_{new}$ represents the random $\{\bm{\theta}, \bm{f}\}$ after the muscle addition).
  The data of $(\bm{\theta}^{data}_{new}, \bm{f}^{data}_{new}, \bm{l}^{data}_{new})$ obtained at this time is $D_{new}$ and the number of data is $N_{new}$.
}%
{%
  まず, それぞれの筋は以下の筋剛性制御\cite{shirai2011stiffness}によって動作する.
  \begin{align}
    \bm{f}^{ref} = \bm{f}_{bias} + k_{stiff}(\bm{l}-\bm{l}^{ref}) \label{eq:msc}
  \end{align}
  $\bm{f}_{bias}$は筋剛性制御のバイアス項, $k_{stiff}$は筋剛性制御の筋剛性係数である.

  筋増加の際, まずは新しくデータを取得する必要がある.
  この際, 新しく追加された筋については$(\bm{\theta}, \bm{f}, \bm{l})$の関係が分かっていないため, この筋のみ他の筋とは別の駆動を行いデータを取る.
  ここで, \equref{eq:msc}における$k_{stiff}$を新しい筋のみ0とし, 指令筋長への追従を行わない.
  その代わり, $\bm{f}^{bias}$をランダムに指定する.
  元からある筋に対しては, $(\bm{\theta}^{rand}_{new}, \bm{f}^{rand}_{new}, \bm{0}, \begin{pmatrix}1&1&0\end{pmatrix}^{T})$を$\bm{h}_{new}$に入力し, 得られた$\bm{l}^{ref}$を実機に送る($\{\bm{\theta}, \bm{f}\}^{rand}_{new}$は筋増加後のランダムな$\{\bm{\theta}, \bm{f}\}$を表す).
  このときに得られた$(\bm{\theta}^{data}_{new}, \bm{f}^{data}_{new}, \bm{l}^{data}_{new})$のデータを$D_{new}$, そのデータ数を$N_{new}$とする.
}%

\subsection{Retraining of Body Schema} \label{subsec:retraining}
\switchlanguage%
{%
  Finally, by simultaneously using $\bm{h}_{old}$ and the obtained data $D_{new}$, $\bm{h}_{new}$ can be efficiently relearned even with a small number of motion data.
  In the case where $D_{new}$ is a relatively small number of data, if $\bm{h}_{new}$ is learned using only these data, it will be overfitted to only them and the information of $\bm{h}_{old}$ will be forgotten.
  On the other hand, since the structure of the entire network changes with the addition of new muscles, the information in $\bm{h}_{old}$ is not completely correct for $\bm{h}_{new}$, though it can be used as a reference.
  Therefore, in this study, we define the loss function as follows to learn $\bm{h}_{new}$.
  \begin{align}
    L &= L_{new} + w_{loss}L_{old}\\
    L_{new} &= ||\bm{\theta}^{pred}_{new}-\bm{\theta}^{data}_{new}||_{2}+||\bm{f}^{pred}_{new}-\bm{f}^{data}_{new}||_{2}+||\bm{l}^{pred}_{new}-\bm{l}^{data}_{new}||_{2}\\
    L_{old} &= ||\bm{\theta}^{pred}_{old'}-\bm{\theta}^{rand}_{old'}||_{2}+||\bm{r}\otimes(\bm{f}^{pred}_{old'}-\bm{f}^{rand}_{old'})||_{2}\nonumber\\
    &\;\;\;\;\;\;\;\;\;\;\;\;\;\;\;\;\;\;\;\;\;\;\;\;\;\;\;\;\;\;\;\;\;\;\;\;+||\bm{r}\otimes(\bm{l}^{pred}_{old'}-\bm{l}^{rand}_{old'})||_{2}
  \end{align}
  where $L_{new}$ is the loss for $D_{new}$, $L_{old}$ is the loss for $\bm{h}_{old}$, and $w_{loss}$ is a coefficient of the weight for the loss $L_{old}$.
  $||\bullet||_{2}$ is L2 norm, $\{\bm{\theta}, \bm{f}, \bm{l}\}^{pred}_{new}$ is the value predicted when inputting $\{\bm{\theta}, \bm{f}, \bm{l}\}^{data}_{new}$ into $\bm{h}_{new}$, and $\{\bm{\theta}, \bm{f}, \bm{l}\}^{pred}_{old'}$ is the value predicted when inputting $\{\bm{\theta}, \bm{f}, \bm{l}\}^{rand}_{old'}$ into $\bm{h}_{new}$.
  $\bm{r}$ is a mask value that indicates 1 for the original muscle and 0 for the newly added muscle, and $\otimes$ expresses element-wise product.
  Here, for $L_{old}$, we need to calculate $\{\bm{\theta}, \bm{f}, \bm{l}\}^{rand}_{old'}$.
  First, we prepare a random value of $\{\bm{\theta}, \bm{f}\}^{rand}_{old}$ as an input to $\bm{h}_{old}$, and calculate $\bm{l}^{rand}_{old}$ from $\bm{h}_{old}(\bm{\theta}^{rand}_{old}, \bm{f}^{rand}_{old}, \bm{0}, \begin{pmatrix}1&1&0\end{pmatrix}^{T})$.
  Next, in order to use this data as input to $\bm{h}_{new}$, we create $\{\bm{\theta}, \bm{f}, \bm{l}\}^{rand}_{old'}$ with 0 for the newly added muscles.
  Since the information on the newly added muscles is not accurate, we apply the mask $\bm{r}$ and consider this as a loss.
  Note that $\bm{\theta}^{rand}_{old'}=\bm{\theta}^{rand}_{old}$.
  While this process is similar to Network Distillation \cite{masana2020incremental}, it differs in that dimensions are added to the inputs and outputs of the network and $\bm{h}_{old}$ is not necessarily correct for $\bm{h}_{new}$.

  In this study, we compare the following three cases.
  \begin{enumerate}
    \renewcommand{\labelenumi}{(\roman{enumi})}
    \item $w_{loss}=0$
    \item $w_{loss}=1.0$
    \item $w_{loss}=1.0-e/N_{epoch}$
  \end{enumerate}
  where $e$ is the current number of epochs and $N_{epoch}$ is the total number of epochs in the training.
  The hypothesis is that (i) will overfit to $D_{new}$ and forget the information of $\bm{h}_{old}$, (ii) will always contain information of $\bm{h}_{old}$ which is not correct for $\bm{h}_{new}$, and (iii) will be the best way to relearn $\bm{h}_{new}$ from a small number of motion data.
}%
{%
  最後に, $\bm{h}_{old}$と得られたデータ$D_{new}$を同時に使うことで, 少数のデータでも効率的に$\bm{h}_{new}$を再学習する.
  $D_{new}$が比較的少数のデータの場合, これらのデータだけで$\bm{h}_{new}$を学習してしまうとそのデータだけに過学習してしまい, $\bm{h}_{old}$の情報を忘却してしまう.
  一方, 筋が追加されることによってネットワーク全体の構造も変化するため, $\bm{h}_{old}$の情報は参考にはなるものの$\bm{h}_{new}$として正しくはない.
  そこで, 本研究では以下のように損失関数を定義して$\bm{h}_{new}$の学習を行う.
  \begin{align}
    L &= L_{new} + w_{loss}L_{old}\\
    L_{new} &= ||\bm{\theta}^{pred}_{new}-\bm{\theta}^{data}_{new}||_{2}+||\bm{f}^{pred}_{new}-\bm{f}^{data}_{new}||_{2}+||\bm{l}^{pred}_{new}-\bm{l}^{data}_{new}||_{2}\\
    L_{old} &= ||\bm{\theta}^{pred}_{old'}-\bm{\theta}^{rand}_{old'}||_{2}+||\bm{r}\otimes(\bm{f}^{pred}_{old'}-\bm{f}^{rand}_{old'})||_{2}\nonumber\\
    &\;\;\;\;\;\;\;\;\;\;\;\;\;\;\;\;\;\;\;\;\;\;\;\;\;\;\;\;\;\;\;\;\;\;\;\;+||\bm{r}\otimes(\bm{l}^{pred}_{old'}-\bm{l}^{rand}_{old'})||_{2}
  \end{align}
  ここで, $L_{new}$は$D_{new}$に関する損失, $L_{old}$は$\bm{h}_{old}$に関する損失, $w_{loss}$は損失に関する重みの係数を表す.
  $\{\bm{\theta}, \bm{f}, \bm{l}\}^{pred}_{new}$は$\bm{h}_{new}$に$\{\bm{\theta}, \bm{f}, \bm{l}\}^{data}_{new}$を入力した際に予測された値, $\{\bm{\theta}, \bm{f}, \bm{l}\}^{pred}_{old'}$は$\bm{h}_{new}$に$\{\bm{\theta}, \bm{f}, \bm{l}\}^{rand}_{old'}$を入力した際に予測された値を表す.
  $\bm{r}$は元からある筋に対しては1, 新しく追加された筋に対しては0を示すマスク値であり, $\otimes$は要素ごとの積を表す.
  ここで, $L_{old}$について, $\{\bm{\theta}, \bm{f}, \bm{l}\}^{rand}_{old'}$を求める必要がある.
  まず$\bm{h}_{old}$への入力としてランダムな$\{\bm{\theta}, \bm{f}\}^{rand}_{old}$を用意し, $\bm{h}_{old}(\bm{\theta}^{rand}_{old}, \bm{f}^{rand}_{old}, \bm{0}, \begin{pmatrix}1&1&0\end{pmatrix}^{T})$から$\bm{l}^{rand}_{old}$を計算する.
  次に, このデータを$\bm{h}_{new}$への入力とするため, 新しく追加された筋については0を付け加えた$\{\bm{\theta}, \bm{f}, \bm{l}\}^{rand}_{old'}$を作成する.
  新しく追加された筋の情報は正確ではないため, マスク$\bm{r}$をかけ, これを損失とする.
  なお、$\bm{\theta}^{rand}_{old'}=\bm{\theta}^{rand}_{old}$である.
  本研究はDistillation \cite{masana2020incremental}と似た考え方である一方, ネットワークの入出力に対して次元が追加される点・$\bm{h}_{old}$が$\bm{h}_{new}$に対して必ず正しいわけではない点が異なる.

  本研究では以下の3つのケースについて比較・考察も行う.
  \begin{enumerate}
    \renewcommand{\labelenumi}{(\roman{enumi})}
    \item $w_{loss}=0$
    \item $w_{loss}=1.0$
    \item $w_{loss}=1.0-e/N_{epoch}$
  \end{enumerate}
  ここで, $e$は現在のエポック数, $N_{epoch}$は学習全体のエポック数を表す.
  仮説として, (i)は過学習し$\bm{h}_{old}$の情報を忘却してしまい, (ii)は$\bm{h}_{new}$としては正しくない$\bm{h}_{old}$の情報が常に入り続けてしまい, (iii)が最も良く少数のデータから$\bm{h}_{new}$を再学習することができると考えられる.
}%

\begin{figure}[t]
  \centering
  \includegraphics[width=0.9\columnwidth]{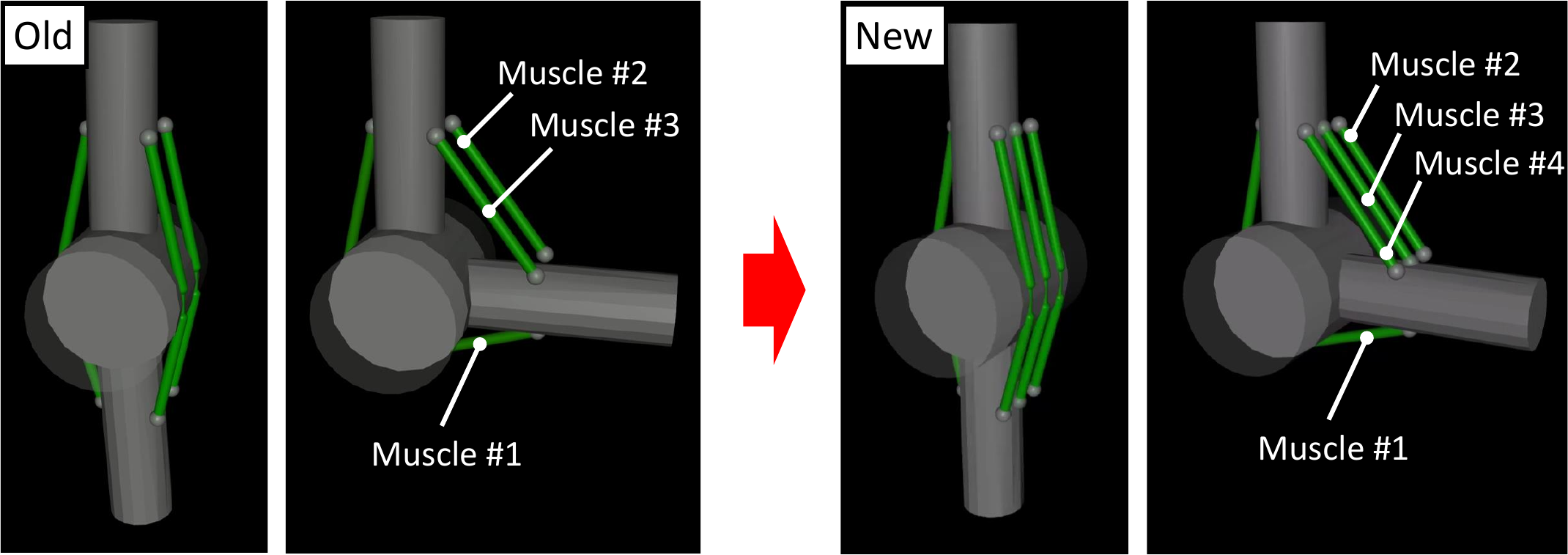}
  \vspace{-1.0ex}
  \caption{The old and new designs of muscle arrangement for the experiment of a simple 1-DOF tendon-driven robot simulation.}
  \label{figure:sim-exp}
  \vspace{-1.0ex}
\end{figure}

\begin{figure}[t]
  \centering
  \includegraphics[width=0.7\columnwidth]{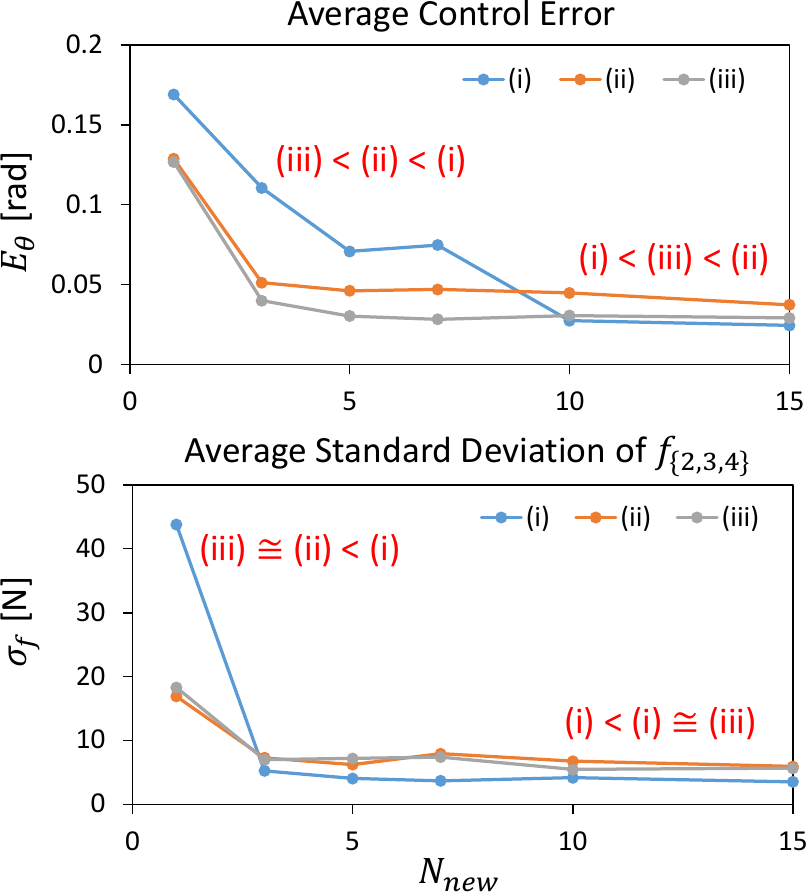}
  \vspace{-1.0ex}
  \caption{The evaluation experiment of a simple 1-DOF tendon-driven robot simulation. The graphs show $E_{\theta}$ (upper) and $\sigma_{f}$ (lower) when changing the number of collected data $N_{new}$, regarding the methods of (i)--(iii).}
  \label{figure:sim-eval}
  \vspace{-3.0ex}
\end{figure}

\section{Experiments} \label{sec:experiments}
\subsection{Simple 1-DOF Tendon-driven Robot Simulation} \label{subsec:sim-exp}
\switchlanguage%
{%
  This section describes a simple 1-DOF tendon-driven robot simulation experiment.
  We create a 1-DOF, 3-muscle robot on MuJoCo \cite{todorov2012mujoco} that mimics a human elbow as shown in the left figures of \figref{figure:sim-exp}.
  To make the robot closer to the actual robot, nonlinear elastic elements with similar properties to those of the muscles in Musashi \cite{kawaharazuka2019musashi} are used, the muscles are configured to have frictional loss, and the moment arm of the muscles changes according to the joint angle.
  Two flexor muscles (\#2 and \#3) and one extensor muscle (\#1) are placed.
  This original muscle arrangement is called ``Old'', and the muscle arrangement with one more flexor muscle (\#4) is called ``New''.
  Adaptive body schema learning experiments are conducted using these two arrangements.

  First, we create $\bm{h}_{new}$ from the well-trained $\bm{h}_{old}$ by copying the network weights using the method of \secref{subsec:changed-schema}.
  Next, we collect $D_{new}$ using the method of \secref{subsec:data-collection} and update $\bm{h}_{new}$ by the methods (i)--(iii) of \secref{subsec:retraining}.
  An evaluation experiment is performed on the obtained $\bm{h}_{new}$.
  16 target joint angles $\bm{\theta}^{ref}$ are set ($\bm{\theta}^{ref}$ is generated by going from 0 to 120 deg in 15 deg steps and back to account for hysteresis), control is performed using MAE, and the average control error $E_{\theta}=||\bm{\theta}^{ref}-\bm{\theta}||_{2}$ and the average standard deviation of the muscle tensions of the three flexor muscles (\#2--\#4) $\sigma_{f}$ are evaluated.
  Both $E_{\theta}$ and $\sigma_{f}$ should be small.
  We also change the number of data $N_{new}$ to $\{1, 3, 5, 7, 10, 15\}$, and discuss the change in the evaluation value.
  The results are shown in \figref{figure:sim-eval}.
  When $N_{new}$ is small, $E_{\theta}$ is small in the order of (iii)$<$(ii)$<$(i).
  On the other hand, when $N_{new}$ exceeds 10, (i)$<$(iii)$<$(ii).
  For $\sigma_{f}$, (i)$\simeq$(iii)$<$(ii) when $N_{new}$ is extremely small, but (i)$<$(iii)$\simeq$(ii) after that.
  Note that for the case before relearning, $E_{\theta}=0.0214$ and $\sigma_{f}=7.74$.
  Using (i) when $N_{new}=15$, $E_{\theta}=0.0245$ and $\sigma_{f}=3.54$, which means that $E_{\theta}$ is about the same and $\sigma_{f}$ is reduced to less than half, when compared to before the relearning.
  In the case of (i) without weight copying of \secref{subsec:changed-schema}, $E_{\theta}=0.158$ even when $N_{new}=15$, so the weight copying is essential.
}%
{%
  1自由度腱駆動シミュレーション実験について説明する.
  \figref{figure:sim-exp}の左図に示すような人間の肘を模した1自由度3筋のロボットをMuJoCo \cite{todorov2012mujoco}上に作成する.
  実機に近づくようMusashi \cite{kawaharazuka2019musashi}の筋と同様の特性を持つ非線形弾性要素を配置し, 筋に摩擦損失を持たせ, 筋のモーメントアームも関節角度に応じて変化するような構成としている.
  筋配置としては, 屈曲側に2本の筋(\#2, \#3), 伸展側に1本の筋(\#1)を配置している.
  この元々の筋配置をOld, この屈曲側の筋を3本(\#2, \#3, \#4)に増やした筋配置をNewと呼び, 筋増加に伴うモデル学習実験を行う.

  まず, 十分に学習された$\bm{h}_{old}$から, \secref{subsec:changed-schema}の方法で重みのコピーにより$\bm{h}_{new}$を作成する.
  次に, \secref{subsec:data-collection}の方法により$D_{new}$を収集し, \secref{subsec:retraining}の(i)--(iii)の方法で$\bm{h}_{new}$を更新する.
  得られた$\bm{h}_{new}$に対して評価実験を行う.
  指令関節角度$\bm{\theta}^{ref}$を16個定め(ヒステリシスを考慮して0--120 degの間を15 degずつ行って戻る$\bm{\theta}^{ref}$を生成した), MAEを使った制御を行い, この際の制御誤差の平均$E_{\theta}=||\bm{\theta}^{ref}-\bm{\theta}||_{2}$と屈曲側の3つの筋(\#2--\#4)の筋張力の標準偏差の平均$\sigma_{f}$を評価する.
  $E_{\theta}$, $\sigma_{f}$ともに小さくなることが望ましい.
  また, データ数$N_{new}$を$\{1, 3, 5, 7, 10, 15\}$に変化させ, このときの評価値の変化についても考察する.
  その結果を\figref{figure:sim-eval}に示す.
  $E_{\theta}$は, $N_{new}$が小さいときは(iii)$<$(ii)$<$(i)の順で小さい.
  一方, $N_{new}$が10を超えたところで(i)$<$(iii)$<$(ii)となった.
  $\sigma_{f}$は, $N_{new}$が極端に少ないときは(i)$\simeq$(iii)$<$(ii)であるが, それ以降は(i)$<$(iii)$\simeq$(ii)となった.
  なお, 再学習前については, $E_{\theta}=0.0214$, $\sigma_{f}=7.74$であった.
  $N_{new}=15$において(i)は$E_{\theta}=0.0245$, $\sigma_{f}=3.54$であり, $E_{\theta}$は同程度, $\sigma_{f}$は半分以下まで下げることができていた.
  また, \secref{subsec:changed-schema}の重みコピーを行わない(i)の場合は$N_{new}=15$でも$E_{\theta}=0.158$であり, 重みコピーは必須であった.
}%

\begin{figure}[t]
  \centering
  \includegraphics[width=0.9\columnwidth]{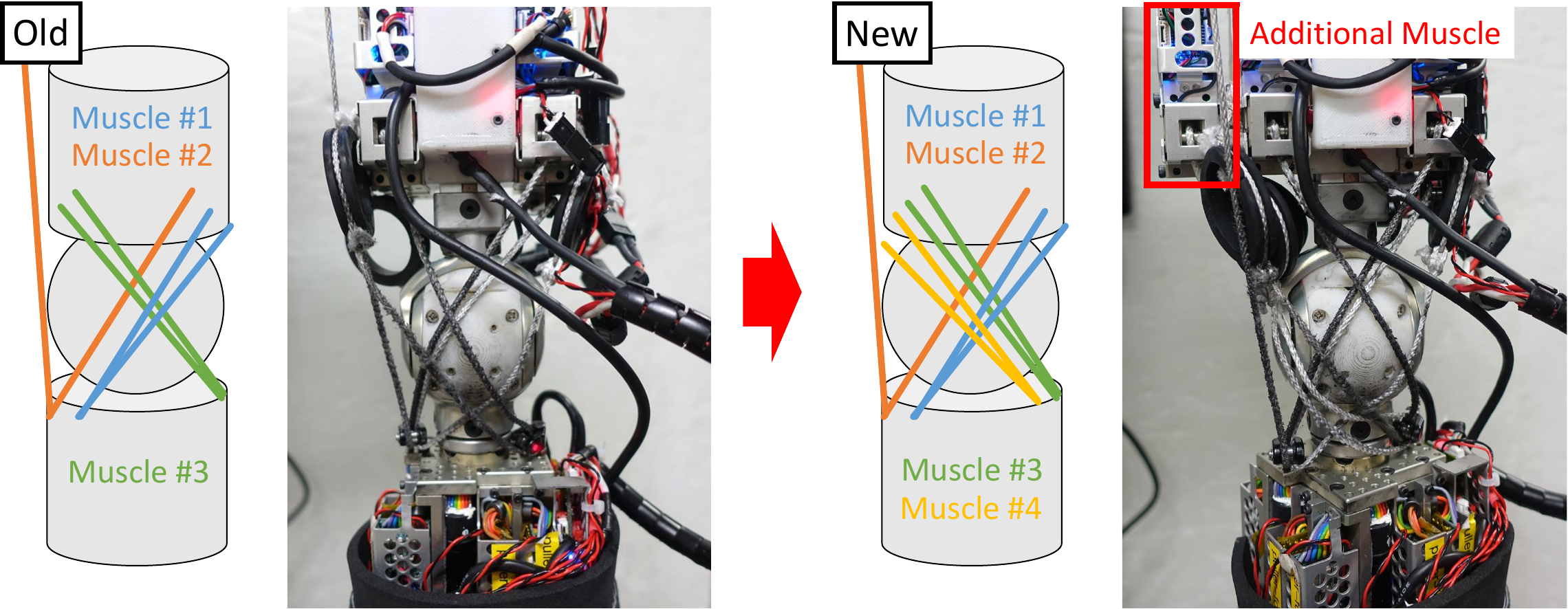}
  \vspace{-1.0ex}
  \caption{The old and new designs of muscle arrangement for the experiment of the left arm of the musculoskeletal humanoid Musashi.}
  \label{figure:act-exp}
  \vspace{-3.0ex}
\end{figure}

\begin{figure*}[t]
  \centering
  \includegraphics[width=1.9\columnwidth]{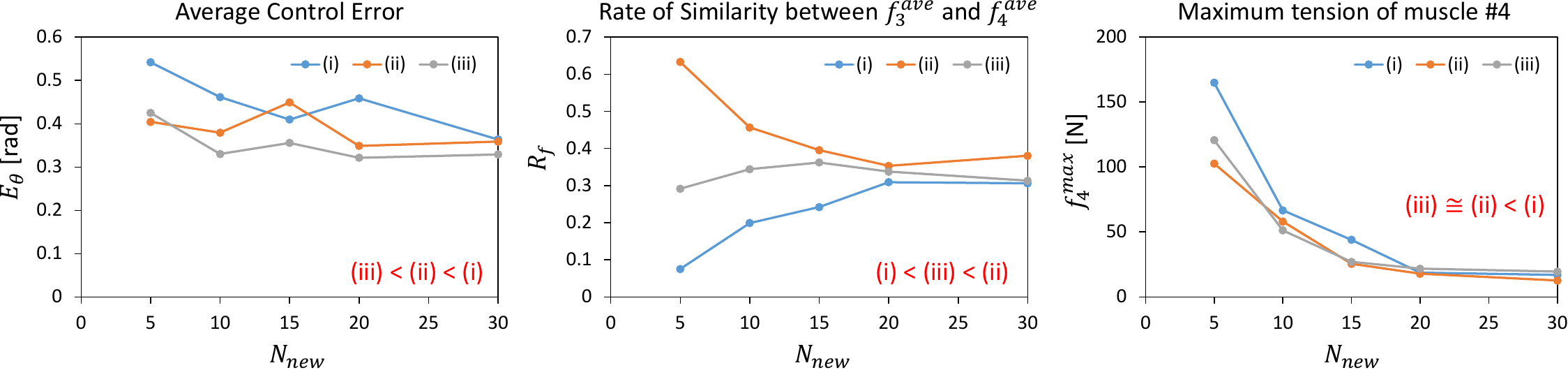}
  \vspace{-1.0ex}
  \caption{The evaluation experiment of the left arm of the musculoskeletal humanoid Musashi. The graphs show $E_{\theta}$ (left), $R_{f}$ (center), and $f^{max}_{4}$ (right) when changing the number of collected data $N_{new}$, regarding the methods of (i)--(iii).}
  \label{figure:act-eval}
  \vspace{-3.0ex}
\end{figure*}

\subsection{The Musculoskeletal Humanoid Musashi} \label{subsec:act-exp}
\switchlanguage%
{%
  Next, we describe experiments using the left arm of the musculoskeletal humanoid Musashi \cite{kawaharazuka2019musashi}.
  We create MAE for 10 muscles with five DOFs in the shoulder and elbow of the left arm and conduct experiments.
  As shown in \figref{figure:act-exp}, three flexor muscles (\#1, \#2, \#3) of the elbow are arranged with the addition of a fourth muscle (\#4).
  We call the original arrangement with 10 muscles ``Old'' and the new arrangement with 11 muscles ``New'', and conduct adaptive body schema learning experiments with the additional muscle.

  We obtain $D_{new}$ in the same way as in \secref{subsec:sim-exp} and update $\bm{h}_{new}$ by the methods of (i)--(iii).
  For the evaluation experiment, we set 10 random target joint angles $\bm{\theta}^{ref}$ and evaluate the average of the control errors $E_{\theta}$ for them.
  It is difficult to evaluate the muscle tension in the same way as for \secref{subsec:sim-exp}, because different moment arms often produce completely different muscle tensions at the same timing.
  Therefore, in this study, we evaluate $E_{f}=|f^{ave}_{3}-f^{ave}_{4}|/(f^{ave}_{3}+f^{ave}_{4})$ and $f^{max}_{4}$.
  Note that $f^{ave}_{\{3, 4\}}$ represents the average muscle tension of muscle \#3 and \#4, $|\bullet|$ represents an absolute value, and $f^{max}_{4}$ represents the maximum muscle tension of muscle \#4 in the evaluation experiment.
  This allows us to know whether $f_{3}$ and $f_{4}$, which have similar roles, have similar muscle tension values throughout the entire evaluation experiment, and whether the newly added muscle \#4 is subjected to an unreasonable force.
  Since the correct ratio of $f^{ave}_{3}$ to $f^{ave}_{4}$ is not known due to the difference in the moment arms, we use the ratio of the magnitudes of $E_{f}$ for (i)--(iii), $R_{f}$, as the evaluation value for $E_{f}$.
  We also change the number of data $N_{new}$ to $\{5, 10, 15, 20, 30\}$, and consider the change in the evaluation value.
  The results are shown in \figref{figure:act-eval}.
  The values of $E_{\theta}$ are generally smaller in the order of (iii)$<$(ii)$<$(i), although some of them are reversed.
  $R_{f}$ always satisfies (i)$<$(iii)$<$(ii) within the range of this experiment, and the difference becomes smaller as $N_{new}$ increases.
  For $f^{max}_{4}$, (iii)$\simeq$(ii)$<$(i), and the difference becomes smaller as $N_{new}$ increases.
  Note that for the case before retraining, $E_{\theta}=0.526$.
  $E_{\theta}=0.329$ using (iii) when $N_{new}=30$, and the error is greatly reduced by the addition of muscles and retraining of body schema.
}%
{%
  次に, 筋骨格ヒューマノイドMusashi \cite{kawaharazuka2019musashi}の左腕を用いた実験について説明する.
  左腕の中でも肩・肘の5自由度10筋についてMAEを作成し実験を行う.
  \figref{figure:act-exp}に示すように肘の屈曲側に備わる3本の筋(\#1, \#2, \#3)に4つめ(\#4)を加えて配置した.
  元々の10本の筋配置をOld, 新しい11本の筋配置をNewと呼び, 筋増加に伴うモデル学習実験を行う.

  \secref{subsec:sim-exp}と同様に$D_{new}$を取得し, (i)--(iii)の方法で$\bm{h}_{new}$を更新する.
  評価実験については, ランダムな指令関節角度$\bm{\theta}^{ref}$を10個定め, これらに対して制御誤差の平均$E_{\theta}$を評価する.
  筋張力については, モーメントアームがそれぞれ異なるため同じタイミングでは全く異なる筋張力が出ることが多いため\secref{subsec:sim-exp}と同様に評価することは難しい.
  そのため, 本研究では$E_{f}=|f^{ave}_{3}-f^{ave}_{4}|/(f^{ave}_{3}+f^{ave}_{4})$と, $f^{max}_{4}$について評価を行う.
  なお, $f^{ave}_{\{3, 4\}}$は評価実験における筋\#3と\#4の平均の筋張力, $f^{max}_{4}$は評価実験における筋\#4の最大の筋張力を表す.
  これにより, 評価実験中における, 役割の近い$f_{3}$と$f_{4}$が全体として近いかどうか, 新しく追加した筋\#4に無理な力がかかっていないかを知ることができる.
  モーメントアームが異なり$f^{ave}_{3}$と$f^{ave}_{4}$の正しい比率は分からないため, $E_{f}$については(i)--(iii)に関する$E_{f}$の大きさの比率$R_{f}$を評価値とする.
  また, データ数$N_{new}$を$\{5, 10, 15, 20, 30\}$に変化させ, このときの評価値の変化についても考察する.
  その結果を\figref{figure:act-eval}に示す.
  $E_{\theta}$は, 一部逆になることがあるものの, 全体的に(iii)$<$(ii)$<$(i)の順で小さい.
  $R_{f}$は, 実験の範囲内では常に(i)$<$(iii)$<$(ii)であり, その差は$N_{new}$が増えるにつれて小さくなっていった.
  $f^{max}_{4}$は, (iii)$\simeq$(ii)$<$(i)であり, その差は$N_{new}$が増えるにつれて小さくなっていった.
  なお, 再学習前については, $E_{\theta}=0.526$であった.
  $N_{new}=30$において(iii)は$E_{\theta}=0.329$であり, 筋の追加と再学習により誤差は大幅に小さくなった.
}%

\begin{figure}[t]
  \centering
  \includegraphics[width=0.9\columnwidth]{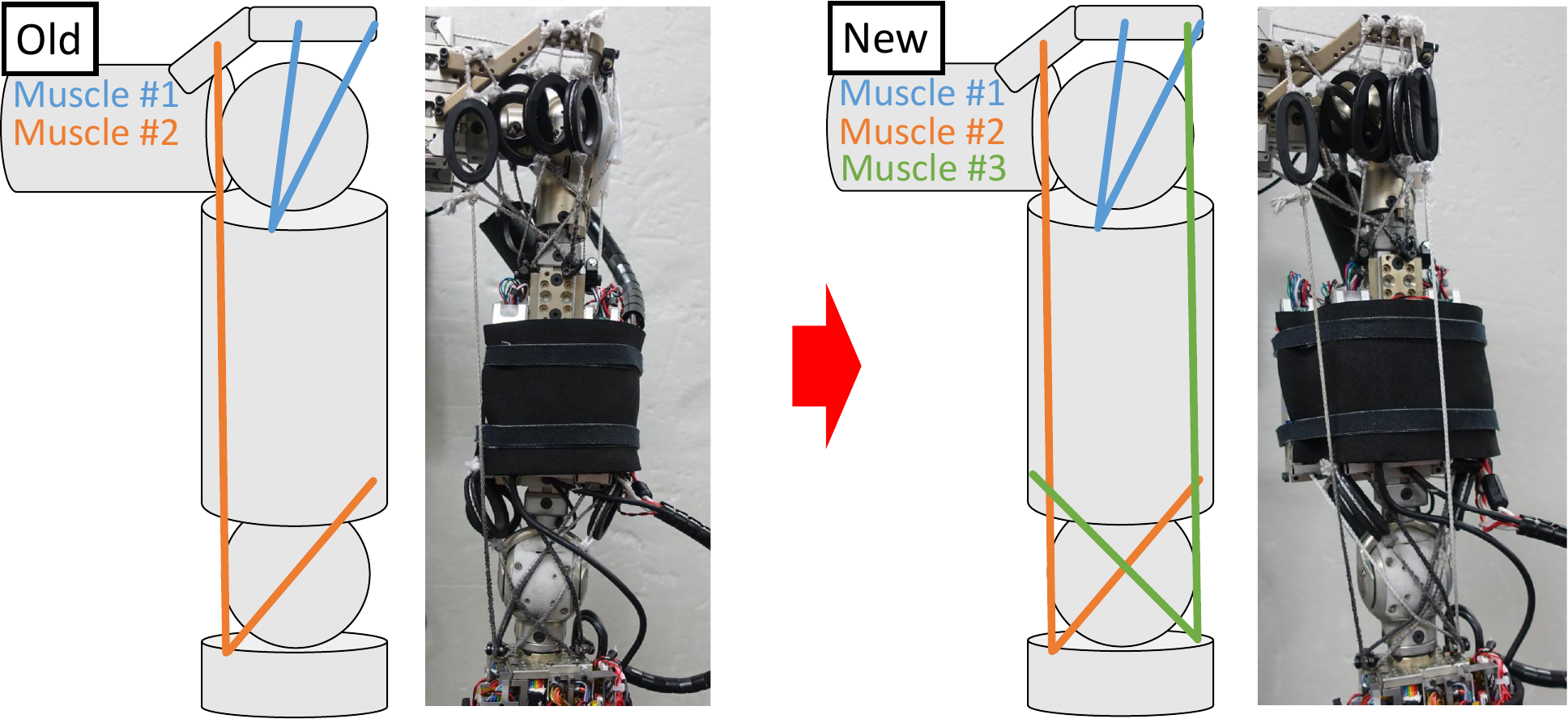}
  \vspace{-1.0ex}
  \caption{The old and new designs of muscle arrangement for a high-load task experiment using Muashi-W.}
  \label{figure:integrated-exp}
  \vspace{-1.0ex}
\end{figure}

\begin{figure}[t]
  \centering
  \includegraphics[width=0.9\columnwidth]{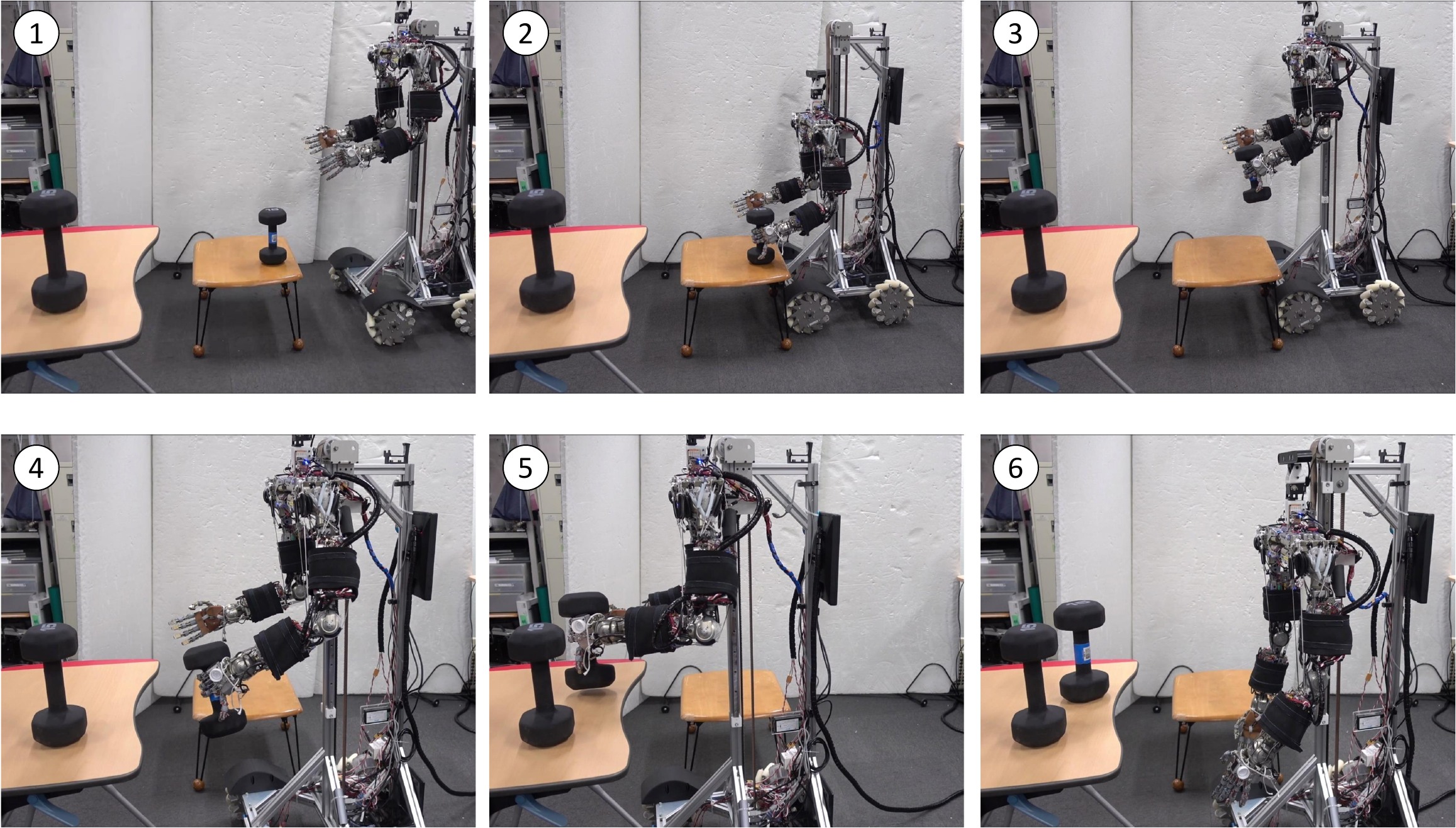}
  \vspace{-1.0ex}
  \caption{The high-load task experiment using Muashi-W.}
  \label{figure:integrated-exp2}
  \vspace{-3.0ex}
\end{figure}

\begin{figure}[t]
  \centering
  \includegraphics[width=0.7\columnwidth]{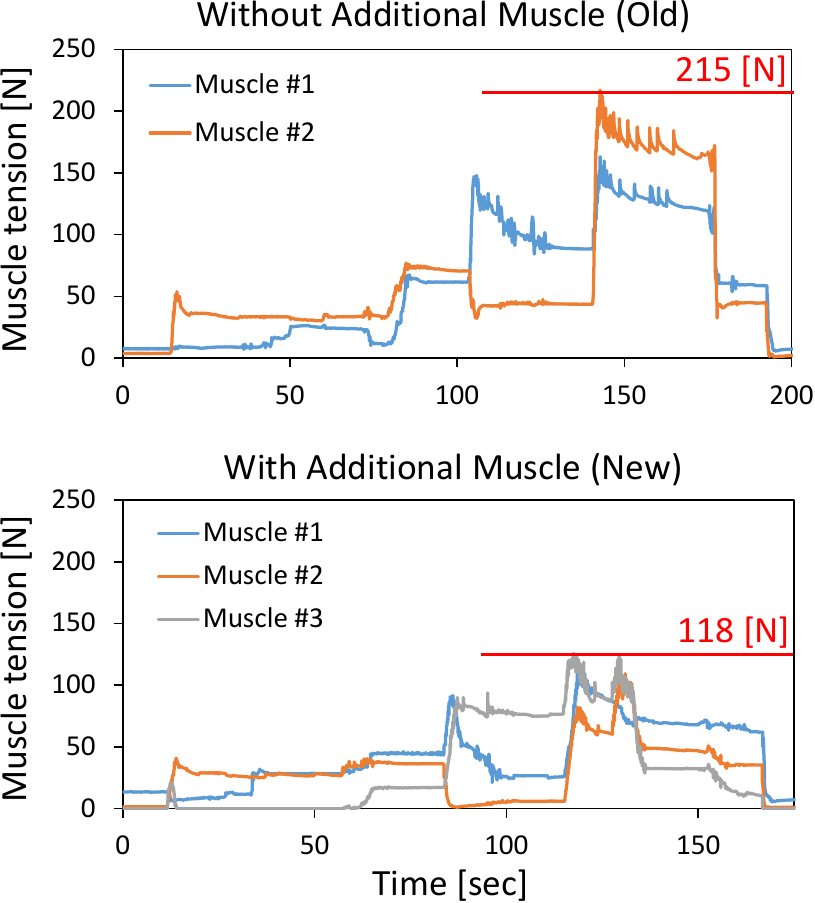}
  \vspace{-1.0ex}
  \caption{The evaluation experiment of the high-load task. The graphs show the transition of muscle tensions related to the movement of the shoulder pitch joint, when using hardware without an additional muscle (upper) and with an additional muscle \#3 (lower).}
  \label{figure:integrated-eval}
  \vspace{-3.0ex}
\end{figure}

\subsection{High-load Task Experiment}
\switchlanguage%
{%
  Finally, a task with high load is performed to demonstrate the effectiveness of the muscle addition.
  In this experiment, we use Musashi-W, which has the same dual arms with the musculoskeletal humanoid Musashi \cite{kawaharazuka2019musashi}, but with a mecanum wheeled base and an additional slider for lifting and lowering.
  The muscles directly involved in the pitch joint of the shoulder (\#1, \#2) are extracted and shown in the left figure of \figref{figure:integrated-exp}.
  We add muscle \#3 to the above and relearn $\bm{h}_{new}$ as in previous experiments.
  In this case, the number of data is $N_{new}=17$, and the relearning is performed using the method (iii), which can achieve both small control error and active use of the additional muscles based on the experiments in the previous section.
  We conducted an experiment in which a dumbbell with about 6.8 kg weight was lifted from a lower table and placed on a high table before and after adding muscles and relearning the body schema (\figref{figure:integrated-exp2}).
  The muscle tension transitions of the muscles directly involved in the pitch joint of the shoulder, which exert the most force during this experiment, are shown in \figref{figure:integrated-eval}.
  Note that although the commanded joint angles are the same, the execution time is different because the wheeled base is operated by a human.
  Before the addition of the muscle, a maximum force of 215 N was applied to the muscle tension, while after the addition of the muscle, the maximum force was 118 N, indicating that the muscle tension was significantly reduced.
  Also, after the addition of the muscle, the muscle tension is uniformly distributed among muscle \#1 -- \#3 at the peak, indicating that the additional muscle is correctly used by the relearning of body schema.
}%
{%
  最後に, 高負荷のかかるタスクを行い本研究の筋追加の有効性を示す.
  本実験では, 筋骨格ヒューマノイドMusashi \cite{kawaharazuka2019musashi}と同じ双腕にメカナム台車と昇降自由度が追加されたMusashi-Wを用いて実験を行う.
  \figref{figure:integrated-exp}の左図に肩のピッチ自由度に直接関与する筋(\#1, \#2)を抜き出し示す.
  これに筋\#3を追加し, これまでと同様に$\bm{h}_{new}$の再学習を行う.
  この際, データ数は$N_{new}=17$であり, 前節までの実験から制御誤差の小ささと追加した筋の積極的利用を両立できる(iii)の方法を使って再学習を行った.
  筋を追加する前と追加し再学習を行った後について, 6.8 kgの重りを持ち上げて低い台から高い台の方へ片付ける実験を行った(\figref{figure:integrated-exp2}).
  この際に最も力のかかる肩のピッチ自由度に直接関与する筋の筋張力遷移を\figref{figure:integrated-eval}に示す.
  送っている関節角度等は同一のものだが, 人間がそのタイミングや台車の操作を行っているため実行時間は異なることに注意されたい.
  筋追加前は筋張力に最大で215 Nの力がかかっているのに対して, 筋追加後は最大で118 Nと, 大幅に筋張力を削減できていることがわかる.
  また, 筋追加後はピーク時には筋\#1--\#3の間で筋張力が一様に分散されており, 再学習によって追加した筋を正しく利用できていることがわかる.
}%

\section{Discussion} \label{sec:discussion}
\switchlanguage%
{%

  In the simulation experiments, we were able to understand the overall characteristics of our method.
  Regarding the control error, the method (iii), in which the information about $\bm{h}_{old}$ gradually decays, is the most accurate when the number of obtained data is small.
  The method (ii), which always uses the information of $\bm{h}_{old}$, is not as accurate as (iii), but is more accurate than (i), which uses only newly obtained data.
  On the other hand, when the number of available data is large, there is no need to use the information of $\bm{h}_{old}$, and the accuracy of (i), which learns only from the obtained data, is the highest.
  For the difference in muscle tensions with the same moment arm, the accuracy of (i) is significantly lower when the number of obtained data is extremely small, while the accuracy of (i) is the highest when there is a certain number of data.
  The information of $\bm{h}_{old}$ may have a slight effect of preventing the active use of the additional muscles.
  Overall, all methods reduced the control error compared to before the relearning and worked to reduce the load on other muscles by using the added muscles.
  In addition, the control error is extremely large when the network weights are not copied, indicating that copying the weights is essential.

  Next, in the actual robot experiment, we obtained the same results as in the simulation experiment.
  A certain number of data would be considered as a smaller number of data in the actual experiment compared to the simulation because the actual experiment deals with five DOFs joints while the simulation deals with one DOF joint.
  The control error is (iii)$<$(ii)$<$(i) throughout, and the characteristics are consistent with the state of $N_{new}<10$ in the simulation.
  The muscle tension of the newly added muscle becomes close to that of the muscle with a similar role by the relearning, and the accuracy is (i)$<$(iii)$<$(ii), which means that the properties are consistent with the state of $N_{new}>1$ in the simulation.
  It is also found that regarding the control error and the muscle tension similarity, the differences among the methods (i)--(iii) become smaller as $N_{new}$ becomes larger.
  In addition, when $N_{new}$ is small, it is necessary to be careful because (i) exerts a higher muscle tension for added muscles than (ii) and (iii).
  From these findings, (iii) is better when we want to relearn body schema with a small number of data, and (i) is better when we can collect a large number of training data.

  Finally, we performed a high-load task with the actual robot and verified the effect of adding muscles.
  By adding muscles and performing the relearning of (iii) with a relatively small number of data ($N_{new}=17$), we found that it is possible to significantly reduce the peak muscle tension compared to the case without adding muscles.
  Even with only 17 data points, it was found that muscle tension could be distributed among several muscles with close moment arms, enabling the task to be performed safely.

  Although this study is applied to MAE, it can be used for various network structures by changing the definition of the loss function.
  This research is not limited by the type of the network structure, but proposes a hardware system that enables easy addition of actuators, and a software system that can move the robot by relearning body schema with additional muscles from a small amount of data.
  In the future, we hope that this study will help the development of robots that can efficiently reconstruct and grow their body schema, taking into account the decrease and increase of the number of sensors and actuators.
}%
{%
  本研究の実験結果について考察する.

  シミュレーション実験では, 本手法の全体的な特性を理解することができた.
  制御誤差について, 得られるデータ数が少ない状況においては, $\bm{h}_{old}$に関する情報を徐々に減衰させる(iii)の手法が最も正確である.
  $\bm{h}_{old}$の情報を常に利用する(ii)は(iii)には劣るもの, 新しく得られたデータのみを用いる(i)に比べると精度は高かった.
  一方, 得られるデータ数が多い場合は, わざわざ$\bm{h}_{old}$の情報を利用する必要はなく, 得られたデータのみから学習を行う(i)の精度が最も高かった.
  これに対して共同筋の筋張力の差は, 得られるデータ数が極端に少ない場合は(i)の精度は著しく低い一方で, ある程度のデータ数があれば(ii), (iii)を抜き(i)の精度が最も高かった.
  $\bm{h}_{old}$の情報は追加した筋を積極的に利用することを妨げる効果があると考えられる.
  全体として, どの方法も再学習前よりも制御誤差を減らし, 追加した筋を利用して他の筋の負荷を下げるように働いていることがわかった.
  また, 重みのコピーを行わない場合は極端に制御誤差が大きく, 重みのコピーは必須であることがわかった.

  次に, 実機実験では本シミュレーションの結果と同様の結果を得ることができた.
  実機実験では1自由度のシミュレーションと比べて5自由度の関節を扱っているため, 同様のデータ数ではシミュレーションにおけるデータ数が非常に少ない状態と等価であると考えられる.
  制御誤差は全体を通して(iii)$<$(ii)$<$(i)であり, シミュレーションにおける$N_{new}<10$の状態と特性は一致している.
  新しく追加した筋と, 役割の近い筋の筋張力は再学習により近くなり, その精度は(i)$<$(iii)$<$(ii)であることから, シミュレーションにおける$N_{new}>1$の状態と特性は一致していると考えられる.
  また, (i)--(iii)の制御誤差と筋張力差の違いは$N_{new}$が大きくなるにつれて小さくなっていくことも分かった.
  加えて, $N_{new}$が小さい時は(i)が(ii), (iii)に比べて高い筋張力を発揮してしまうため, 気をつける必要がある.
  これらの知見から, 少数の$N_{new}$で再学習を行いたい場合は(iii)が良く, たくさんの学習データを集めることが可能な場合は(i)の方法が良いことが分かった.

  最後に, 実際にロボットで高負荷のかかるタスクを実行し, この際の筋追加の効果を検証した.
  筋を追加し, $N_{new}=17$という比較的少数のデータで(iii)の再学習を実行することで, 筋を追加しない場合に比べて大幅に筋張力のピークを下げることが可能であることが分かった.
  たった17のデータでもモーメントアームの近い複数の共同筋に筋張力を分配し, 安全にタスクを実行可能になることがわかった.

  本研究はMusculoskeletal AutoEncoderに対して手法を適用したが, 実際には損失関数の定義を変えるのみで, 様々なネットワーク構造に対して利用することができる.
  本研究はネットワーク構造を限定するものではなく, アクチュエータの容易な追加を可能とするハードウェア・これをもとに少数のデータから身体図式を再学習して動いていくロボットシステムの提案である.
  今後, アクチュエータを増やしたり, 減らしたりするロボットにおいて, ネットワーク入出力次元の減少と増加を考慮し, 効率的に身体図式を再構築・成長していくロボット開発の一助になれば幸いである.
}%

\section{CONCLUSION} \label{sec:conclusion}
\switchlanguage%
{%
  In this study, we realized a musculoskeletal humanoid system utilizing the advantages of the  redundant muscle arrangement and easy muscle addition.
  By allowing muscle modules to be connected to each other, it facilitates easy muscle addition and allows muscles to be added according to the task.
  By automatically acquiring motion data and relearning the body schema in response to the changes caused by the additional muscles, it is possible to generate movements that actively use the added muscles.
  By storing the network before the addition of muscles and using it for training, the system can efficiently relearn the body schema even from a small amount of motion data.
  We have successfully applied this system to a simple 1-DOF tendon-driven robot simulation and the left arm of the actual musculoskeletal humanoid Musashi, and demonstrated the effectiveness of this study.
  In the future, we will develop a robot that can change its body structure freely and can learn and grow its body schema.
}%
{%
  本研究では, 筋骨格ヒューマノイドにおける冗長筋駆動と容易な筋追加を活かした動作実現を行った.
  筋モジュール同士の結合を可能とすることで筋の増加を容易にし, タスクに応じた筋追加が可能となる.
  筋追加による身体図式の変化に対して自動でデータを取得, 身体図式を再学習することで, 追加した筋を積極的に使った動作が生成可能になる.
  この際, 筋追加前のネットワークを保存しておき学習時に利用することで, 少数のデータからでも効率的に身体図式を再学習させることが可能である.
  これらの全体システムを1自由度腱駆動シミュレーション, 筋骨格ヒューマノイドMusashiの左腕の実機に適用し, その有用性を示すことに成功した.
  今後, 身体構造を自在に変化可能かつその身体図式を学習・成長していくロボットの開発を進める.
}%

\section*{Acknowledgement}
The authors would like to thank Yuka Moriya for proofreading this manuscript.

{
  \bibliographystyle{IEEEtran}
  \bibliography{main}
}

\end{document}